\documentclass[conference, 9pt]{IEEEtran}
\IEEEoverridecommandlockouts
\usepackage{cite}
\usepackage{amsmath,amssymb,amsfonts}
\usepackage{tabularx}
\usepackage{algorithmic}
\usepackage{graphicx}
\usepackage{subcaption}
\usepackage{textcomp}
\usepackage{xcolor}
\usepackage{hyperref}

\DeclareMathOperator*{\argmax}{argmax}
\DeclareMathOperator*{\argmin}{argmin}

\usepackage{acronym}

\acrodef{IoT}{Internet of Things}
\acrodef{NN}{Neural Network}
\acrodef{EE}{Early Exit}
\acrodef{EENN}{Early Exit Neural Network}
\acrodef{DL}{Deep Learning}
\acrodef{MCU}{Microcontroller}
\acrodef{DoS}{Denial-of-Service}
\acrodef{RNN}{Recurrent Neural Network}
\acrodef{MAC}{Multiply-Accumulate}
\acrodef{BMBF}{German Federal Ministry of Education and Research}
\acrodef{p.p.}{percentage point}

\acrodef{DD}{Difference Detection}
\acrodef{TP}{Temporal Patience}
\acrodef{RDM}{Range-Doppler Map}

\def\BibTeX{{\rm B\kern-.05em{\sc i\kern-.025em b}\kern-.08em
    T\kern-.1667em\lower.7ex\hbox{E}\kern-.125emX}}

\makeatletter
\newcommand{\linebreakand}{%
  \end{@IEEEauthorhalign}
  \hfill\mbox{}\par
  \mbox{}\hfill\begin{@IEEEauthorhalign}
}
\makeatother

\begin{document}

\title{Temporal Patience: Efficient Adaptive Deep Learning for Embedded Radar Data Processing

\thanks{The project on which this report is based was funded by the German Ministry of Education and Research (BMBF) under the project number 16ME0542K. The responsibility for the content of this publication lies with the author.}
}
\author{\IEEEauthorblockN{Max Sponner}
\IEEEauthorblockA{\textit{Development Center} \\
\textit{Infineon Technologies Dresden GmbH \& Co. KG} \\
Dresden, Germany %\\ Max.Sponner@infineon.com
}
\IEEEauthorblockA{\textit{Chair of Processor Design, CfAED}\\
\textit{Technical University of Dresden} \\
Dresden, Germany \\
0000-0002-4830-9440}
\and
\IEEEauthorblockN{Julius Ott}
\IEEEauthorblockA{\textit{Infineon Technologies AG} \\
Neubiberg, Germany %\\ Julius.Ott@infineon.com
}
\IEEEauthorblockA{\textit{Chair for Design Automation} \\
\textit{Technical University of Munich}\\
Munich, Germany \\
0000-0001-8259-3070}
\and
\IEEEauthorblockN{Lorenzo Servadei}
\IEEEauthorblockA{\textit{Chair for Design Automation} \\
\textit{Technical University of Munich}\\
Munich, Germany \\
0000-0003-4322-834X}
\linebreakand
\IEEEauthorblockN{Bernd Waschneck}
\IEEEauthorblockA{\textit{Infineon Technologies AG} \\
Neubiberg, Germany \\
0000-0003-0294-8594
}
\and
\IEEEauthorblockN{Robert Wille}
\IEEEauthorblockA{\textit{Chair for Design Automation} \\
\textit{Technical University of Munich}\\
Munich, Germany \\
0000-0002-4993-7860}
\and
\IEEEauthorblockN{Akash Kumar}
\IEEEauthorblockA{\textit{Chair of Processor Design, CfAED}\\
\textit{Technical University of Dresden} \\
Dresden, Germany \\
0000-0001-7125-1737}
}
\maketitle
\begin{abstract}
  Radar sensors offer power-efficient solutions for always-on smart devices, but processing the data streams on resource-constrained embedded platforms remains challenging.
  This paper presents novel techniques that leverage the temporal correlation present in streaming radar data to enhance the efficiency of Early Exit Neural Networks for Deep Learning inference on embedded devices.
  These networks add additional classifier branches between the architecture's hidden layers that allow for an early termination of the inference if their result is deemed sufficient enough by an at-runtime decision mechanism.
  Our methods enable more informed decisions on when to terminate the inference, reducing computational costs while maintaining a minimal loss of accuracy.
  Our results demonstrate that our techniques save up to 26\,\% of operations per inference over a Single Exit Network and 12\,\% over a confidence-based Early Exit version.
  %This was combined with an additional reorganization of the input data, enabling a total of 70\,\% to 75\,\% fewer operations per inference with an accuracy reduction of just 4.04 percentage points.
  Our proposed techniques work on commodity hardware and can be combined with traditional optimizations, making them accessible for resource-constrained embedded platforms commonly used in smart devices.
  Such efficiency gains enable real-time radar data processing on resource-constrained platforms, allowing for new applications in the context of smart homes, Internet-of-Things, and human–computer interaction.
\end{abstract}


\section{Introduction} \label{introduction}

Integrating radar technology into \ac{IoT} applications and personal computing devices offers advantages over traditional camera-based solutions, including weather and lighting independence and low power consumption while maintaining high-resolution data generation~\cite{choiDeepLearningApproach2022a, servadeiLabelAwareRankedLoss2022, ottUncertaintybasedMetaReinforcementLearning2022, mauroFewShotUserAdaptableRadarBased2023, mauroFewShotUserDefinableRadarBased2022, zhouMMWRadarBasedTechnologies2020, dongSecureMmWaveRadarBasedSpeaker2021b}.

To achieve state-of-the-art prediction performance for radar data processing, \ac{DL} techniques are required~\cite{choiDeepLearningApproach2022a, servadeiLabelAwareRankedLoss2022, mauroFewShotUserAdaptableRadarBased2023}.
However, deploying \ac{DL} workloads on low-powered \acp{MCU} commonly used in \ac{IoT} products presents challenges.
\Acp{EENN}~\cite{pandaConditionalDeepLearning2016, fangFlexDNNInputAdaptiveOnDevice2020, amthorImpatientDNNsDeep2016, huLearningAnytimePredictions2018} provide a potential solution by incorporating additional classifiers between hidden layers, known as \acp{EE}.
\Acp{EENN} terminate the inference when an \ac{EE} provides sufficient results, thus saving computational resources.
The selection of the appropriate classifier is typically based on available compute resources or the input sample.

Existing approaches have limitations that can lead to reduced accuracy or excessive compute resource usage and do not leverage the properties of streaming data resulting in non-optimal decisions~\cite{nguyenDeepNeuralNetworks2015, haqueILFOAdversarialAttack2020, hongPandaNoIt2021}.
This paper introduces two novel techniques for runtime decision-making:
\acl{DD} (\acs{DD}) \acp{EENN} and \acl{TP} (\acs{TP}) \acp{EENN}.
These techniques improve termination decisions by leveraging \ac{EE} output similarity over time to select the best exit for the current sample, thus focusing the similarity metric on the relevant information of the input data, which is based on features extracted by the already executed \ac{NN} layers.
This creates a simple similarity metric that focuses on relevant features for the \ac{NN}'s task.
Additionally, it improves the efficiency by sharing computations between the network inference and the similarity computation and reduces the memory footprint compared to input filtering approaches which need to store the reference input data that is often significantly larger than the classification output vector.
To the best of our knowledge, this is the first paper that leverages the temporal correlation of the sensor data to guide the termination decision process.% in such a way.
%By combining our novel decision mechanism with restructuring the input tensor, we achieved a total reduction of computations per average inference of up to 70\,\% to 75\,\%.
\section{Related Work} \label{related_work}

% The Related Work section reviews prior research in two areas critical to our contribution: processing radar data with \ac{DL} and \acp{EENN}.

The Related Work section examines prior research in two crucial areas: radar data processing with \ac{DL} and \acp{EENN}.

\subsection{Radar-Processing Applications}\label{related_radar}

\begin{figure*}[ht]
\centering
\fontsize{10}{12}\selectfont
\begin{subfigure}{0.2\textwidth}
\fontsize{10}{12}\selectfont
\includegraphics[width=\linewidth]{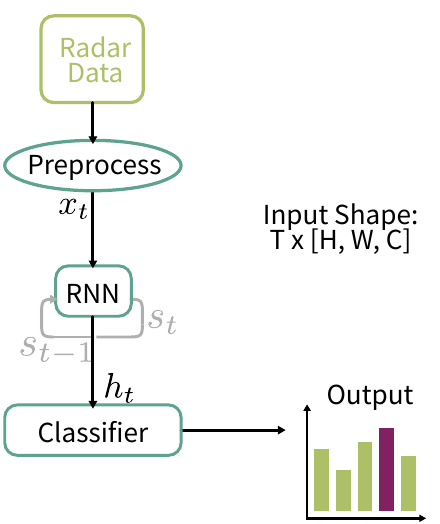}
\caption{Many solutions leverage the temporal correlation of the input to improve the prediction accuracy by using \acp{RNN}.}
\label{fig:rnn}
\end{subfigure}
\hfill
\begin{subfigure}{0.2\textwidth}
\fontsize{10}{12}\selectfont
\includegraphics[width=\linewidth]{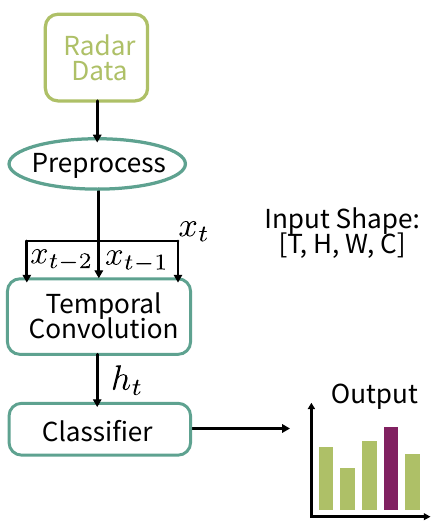}
\caption{An alternative to \ac{RNN}-based architectures is temporal convolutional layers that operate on frames from multiple points in time (using windowing).}
\label{fig:temp_conv}
\end{subfigure}
\hfill
\begin{subfigure}{0.2\textwidth}
\fontsize{10}{12}\selectfont
\includegraphics[width=\linewidth]{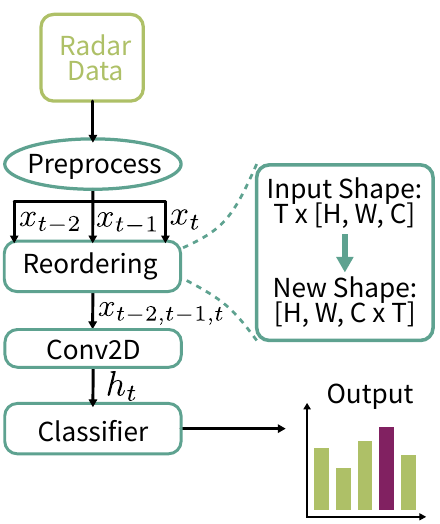}
\caption{Our method rearranges the windowed data by combining the antenna and time dimensions into a single axis to enable the usage of cheaper layer types.}
\label{fig:reorder}
\end{subfigure}
\hfill
\caption{Different approaches to include temporal information into the neural processing network. T are the timesteps, H and W are the dimensions of the feature maps, and C is the number of channels, which equals the number of receiving antennas if Range-Doppler Maps are processed.}
\end{figure*}\noindent

NN-based radar data processing has been explored for various applications, including people-counting~\cite{mauroFewShotUserAdaptableRadarBased2023, servadeiLabelAwareRankedLoss2022, choiDeepLearningApproach2022a}, activity~\cite{wangHumanMotionRecognition2019} and gesture recognition~\cite{mauroFewShotUserDefinableRadarBased2022}.
Specialized algorithms, such as \acp{RNN}~\cite{wangHumanMotionRecognition2019, choiShortRangeRadarBased2019} (see Fig. \ref{fig:rnn}) and Temporal Convolutions~\cite{schererTinyRadarNNCombiningSpatial2021, hazraShortRangeRadarBasedGesture2019} (see Fig.~\ref{fig:temp_conv}) are used to leverage temporal information for enhanced prediction quality.
However, these methods lack support in Embedded Deep Learning toolchains\footnote{\href{https://www.tensorflow.org/lite/microcontrollers}{TensorFlow Lite for Microcontroller}, \href{https://tvm.apache.org/docs/topic/microtvm/index.html}{microTVM}, and \href{https://ai.facebook.com/tools/glow/}{Glow} enable data scientists to compile and optimize their \acp{NN} to be executed on embedded \acp{MCU} without the need for deeper knowledge on these embedded platforms.} or impose a large resource footprint unsuitable for constrained platforms like embedded \acp{MCU}.
% Processing data from multiple radar antennas over time further complicates deployment on limited-capability MCUs, as individual time frames are processed separately, leading to information loss and the inability to track changes in position or speed.

\subsection{Early Exit Neural Networks}\label{related_early}

\Aclp{EENN} (\acsp{EENN}) improve inference efficiency and speed by incorporating multiple output branches, called \aclp{EE} (\acsp{EE}), at different depths.
\Acp{EENN} allow for early predictions based on stopping criteria.
The most commonly used criteria are budget-based and confidence-based.
Budget-based solutions perform the decision before or during the inference based on the availability of compute resources~\cite{amthorImpatientDNNsDeep2016, huLearningAnytimePredictions2018}.
Confidence-based solutions rely on output vector metrics, such as confidence, score margin, or entropy~\cite{pandaConditionalDeepLearning2016, parkBigLittleDeep2015b}.
Such rule-based solutions have limitations, including overspending on simple inputs, incorrect decisions~\cite{nguyenDeepNeuralNetworks2015, kayaShallowdeepNetworksUnderstanding2019}, and vulnerabilities to \ac{DoS} attacks~\cite{haqueILFOAdversarialAttack2020, hongPandaNoIt2021}.
Advanced termination solutions like policy networks require additional resources and training that result in larger resource footprints than a single-exit model would cause and are therefore a poor fit for constrained devices~\cite{bolukbasiAdaptiveNeuralNetworks2017a, odenaChangingModelBehavior2017}.
Explainable AI offers an alternative by identifying relevant filters and reducing computational footprint~\cite{sabihDyFiPExplainableAIbased2022}.
Another approach employs patience, terminating the inference execution when enough subsequent classifiers agree on their output.
However, this requires deeper architectures with a large amount of early exit branches~\cite{zhouBERTLosesPatience2020}.
Template-matching is similar to our approach but lacks temporal components and its input similarity calculation would create too much overhead for the \acp{MCU} due to the large radar data samples~\cite{rashidTemplateMatchingBased2022}.

\section{Use Case} \label{use_case}

% We implemented our solutions for a people-counting application.
% It utilizes the \acp{RDM} acquired by a 60\,GHz radar sensor.
% The \acp{RDM} of the last 8 time-steps are combined into one sample and used as an input for the \ac{NN} - covering 0.8 seconds.
% The output of the model is a classification vector, sorting the current window into a class from zero to four (or more) people present in the surveyed area.

% The training and evaluation datasets were acquired internally using an Infineon XENSIV 60-GHz radar sensor\footnote{\href{https://www.infineon.com/cms/en/product/sensor/radar-sensors/radar-sensors-for-iot/60ghz-radar/bgt60tr13c/}{https://www.infineon.com/cms/en/product/sensor/radar-sensors/radar-sensors-for-iot/60ghz-radar/bgt60tr13c/}}.
% %We are not aware of any comparable publically available datasets, as these usually consist of point-cloud data instead of \acp{RDM}.
% The application targets a Cortex-M4F-based \ac{MCU} for the radar data preprocessing as well as the \ac{NN} inference workload.
% The \ac{NN} is converted using TensorFlow Lite for microcontrollers~\cite{tflite_micro}.

%%%%%%%%%%%%%%%%% SHORTER VERSION %%%%%%%%%%%%%%%%% 
%%%%%%%%%%%%%%%%%%%%%%%%%%%%%%%%%%%%%%%%%%%%%%%%%%%

Our solutions were implemented for a people-counting application using a 60\,GHz radar sensor.
The input consists of groups of \acp{RDM} from the sensor, covering the last eight time-steps.
The output is a classification vector representing the number of people present in the surveyed area.

Training and evaluation datasets were acquired using an Infineon XENSIV 60-GHz radar sensor\footnote{Infineon~XENSIV~60GHz-BGT60TR13: \href{https://www.infineon.com/cms/en/product/sensor/radar-sensors/radar-sensors-for-iot/60ghz-radar/bgt60tr13c/}{https://www.infineon.com/cms/en/product/sensor/radar-sensors/radar-sensors-for-iot/60ghz-radar/bgt60tr13c/}} with an average scene length in the range of thirty seconds.
The application targets a Cortex-M4F-based \ac{MCU} for radar data preprocessing and \ac{NN} inference, utilizing TensorFlow Lite for microcontrollers as deployment toolchain~\cite{tflite_micro}.

\section{Architecture and Training}

The model architecture and training align with the state-of-the-art in \ac{NN} and \ac{EENN}.
We made optimizations to reduce computational requirements, including data reordering and the use of depthwise-separable 2D convolutions.

We employed a data reordering technique to address the challenge of incorporating temporal information while considering resource limitations, as shown in Fig.~\ref{fig:reorder}.
Combining the temporal and antenna axes reduced dimensionality and replaced computationally expensive 3D convolutions with efficient depthwise-separable 2D convolutions~\cite{DBLP:journals/corr/Chollet16a}.
This preserved the ability to process sequence data while reducing computational and memory requirements.

Although the data reordering resulted in a loss of semantic information, the model achieved a training accuracy of 72.5\,\% (compared to 75.33\,\% with 3D convolutions), indicating that the network inferred the lost information during training.
Depthwise-separable convolutions further reduced the resource footprint, achieving a 65.72\,\% reduction in \ac{MAC} operations per inference.

\begin{figure}[ht]
\centering
\fontsize{10}{12}\selectfont
\includegraphics[width=0.9\linewidth]{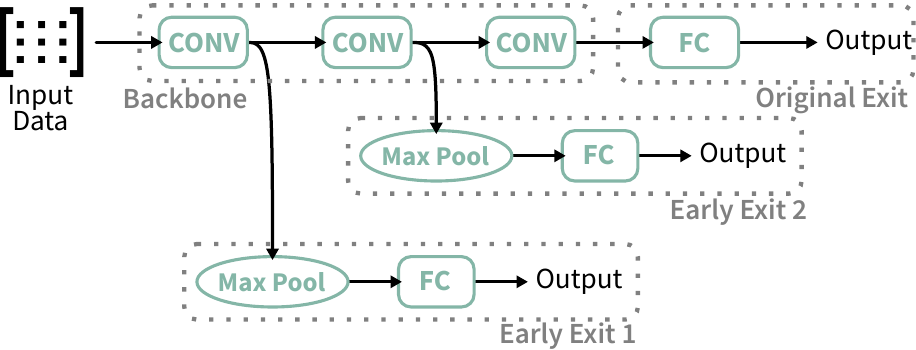}
\caption{The neural network architecture of the base network together with the two added Early Exits.}
\label{fig:architecture}
\end{figure}

The model consists of a backbone and two additional \acp{EE} (see Fig.~\ref{fig:architecture}).
The backbone comprises three convolutional blocks followed by the final classifier, while the \acp{EE} incorporate additional pooling layers to reduce their computational footprint.

Training followed established practices by simultaneously fitting all classifiers and the backbone using backpropagation.

\begin{figure*}[ht]
\centering
\begin{subfigure}{0.23\textwidth} % Adjust the width as needed
\centering
\includegraphics[width=\linewidth]{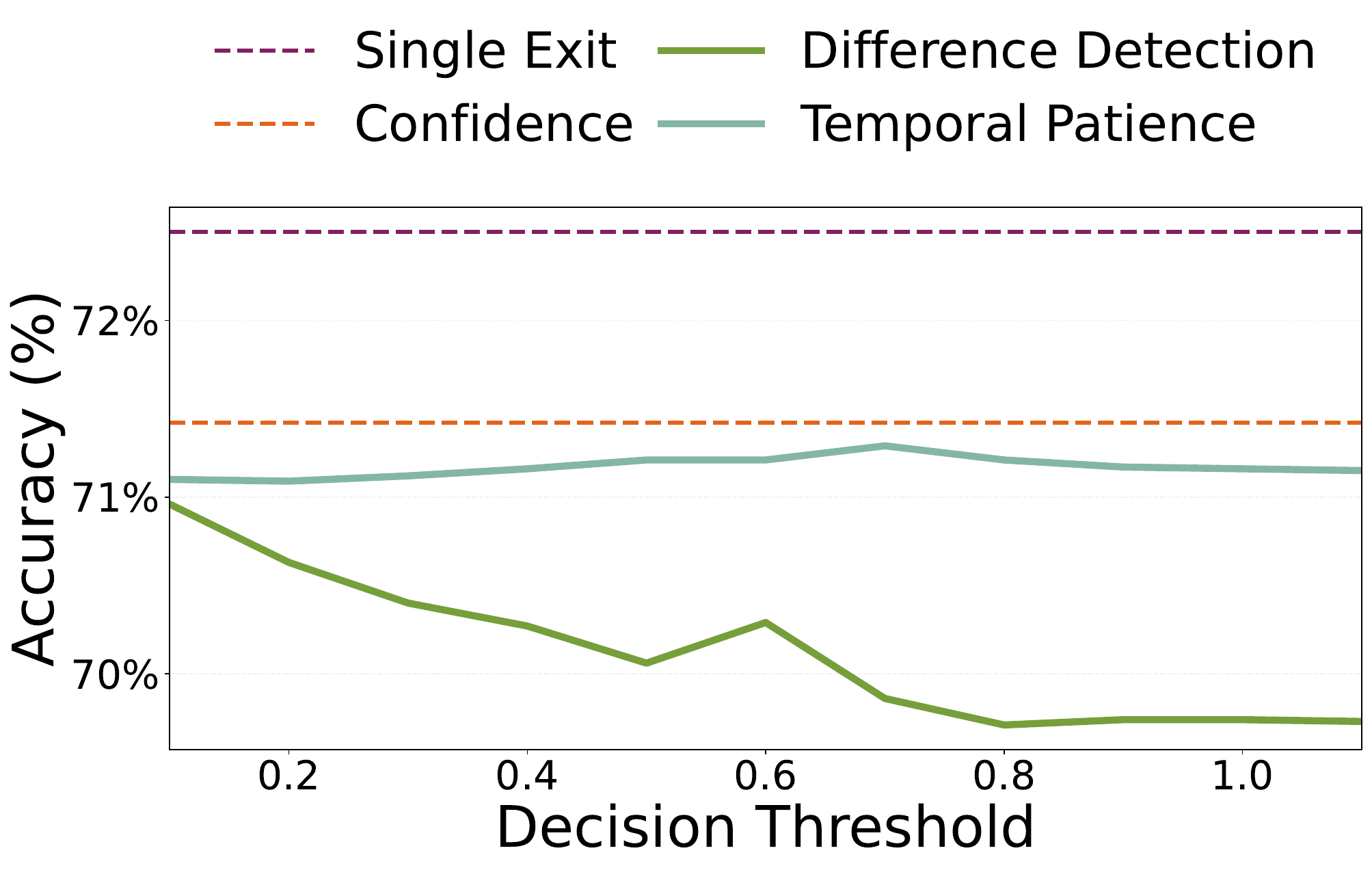}
\caption{Test Set Accuracy across the threshold configurations for all methods.}
\label{fig:acc}
\end{subfigure}\hfill
\begin{subfigure}{0.23\textwidth} % Adjust the width as needed
\centering
\includegraphics[width=\linewidth]{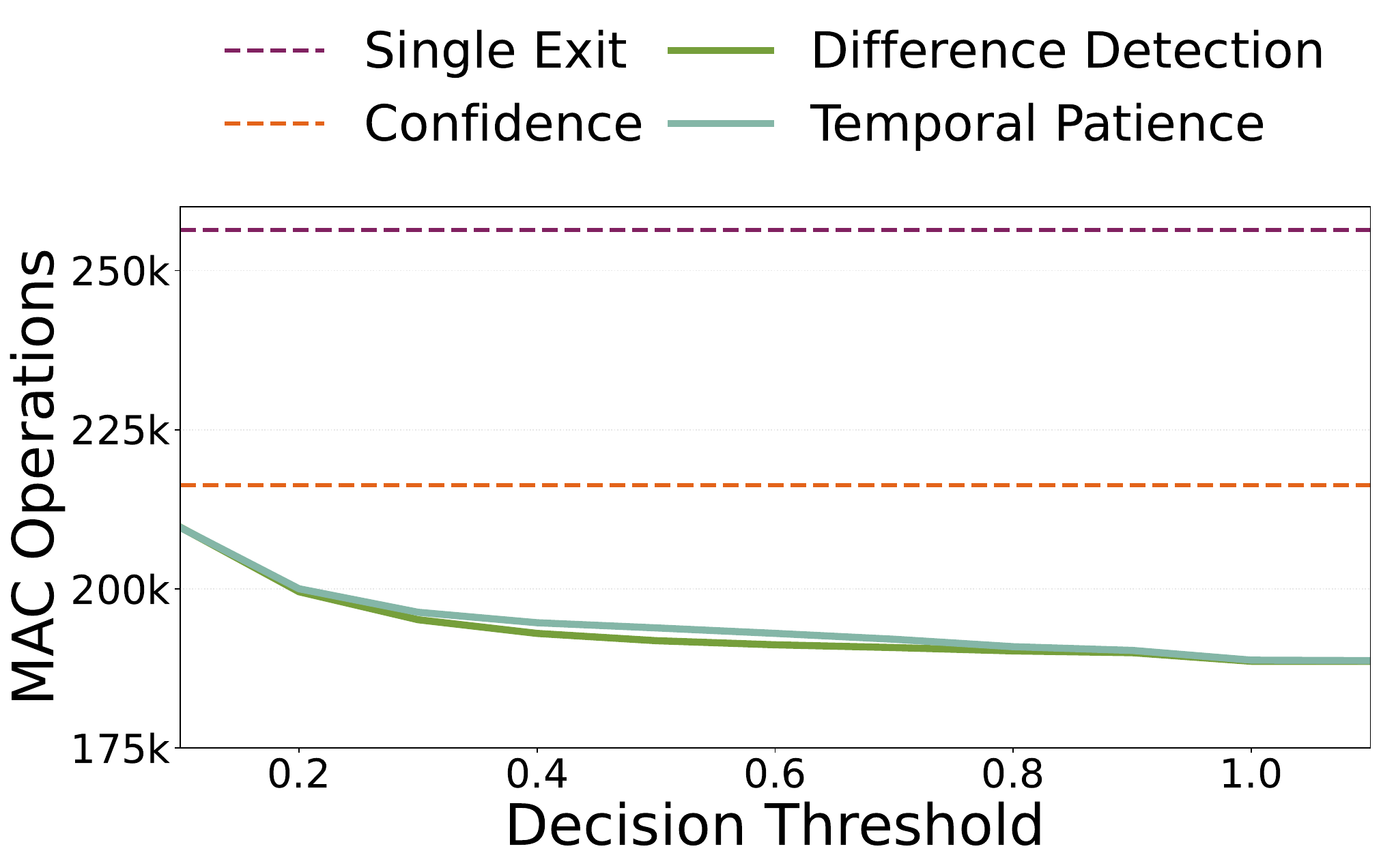}
\caption{Average MAC ops across the threshold configurations for all methods.}
\label{fig:macs}
\end{subfigure}\hfill
\begin{subfigure}{0.23\textwidth} % Adjust the width as needed
\centering
\includegraphics[width=\linewidth]{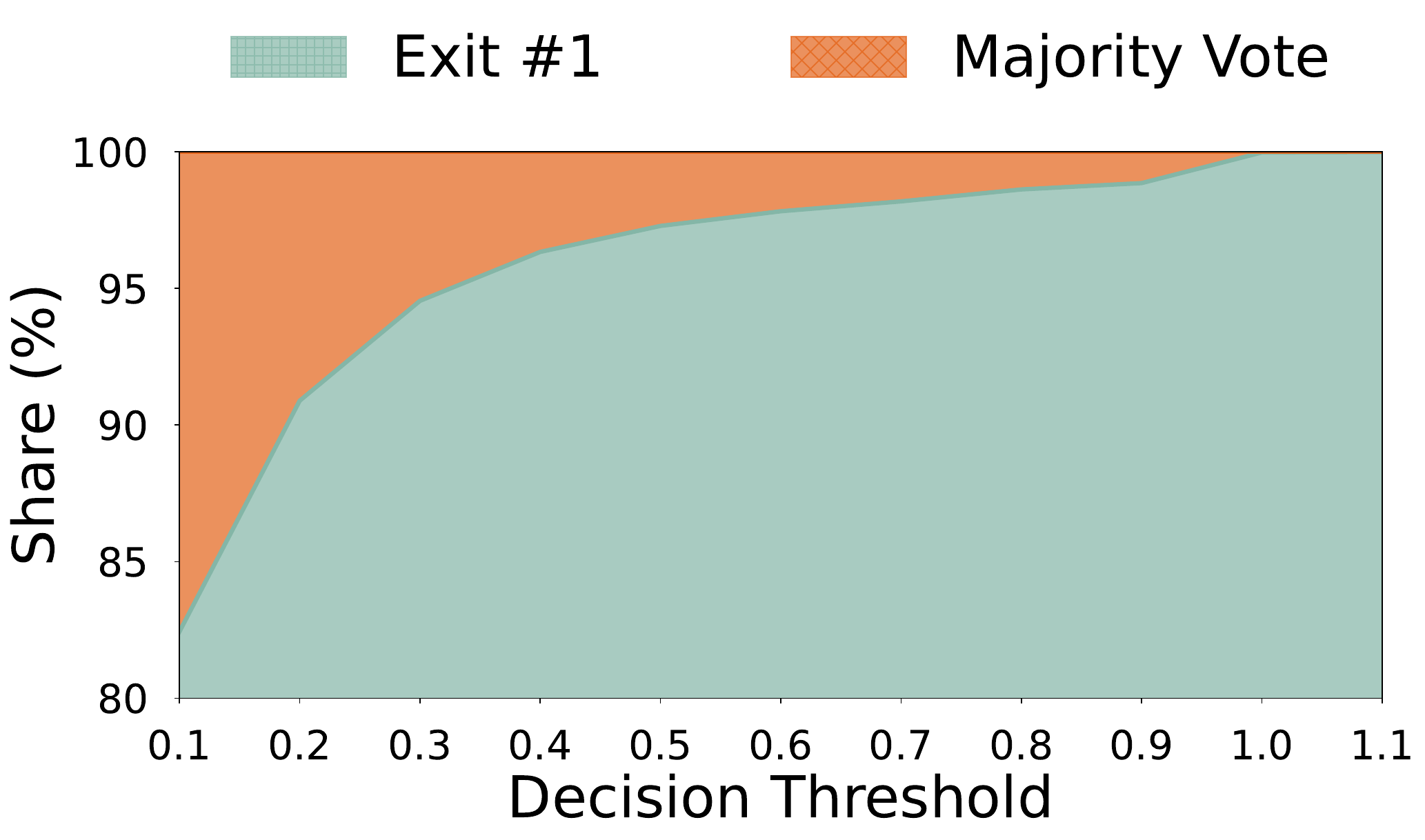}
\caption{Share of used classifiers for \ac{DD} \acp{EENN}.}
\label{fig:share_diff_detect}
\end{subfigure}\hfill
\begin{subfigure}{0.23\textwidth} % Adjust the width as needed
\centering
\includegraphics[width=\linewidth]{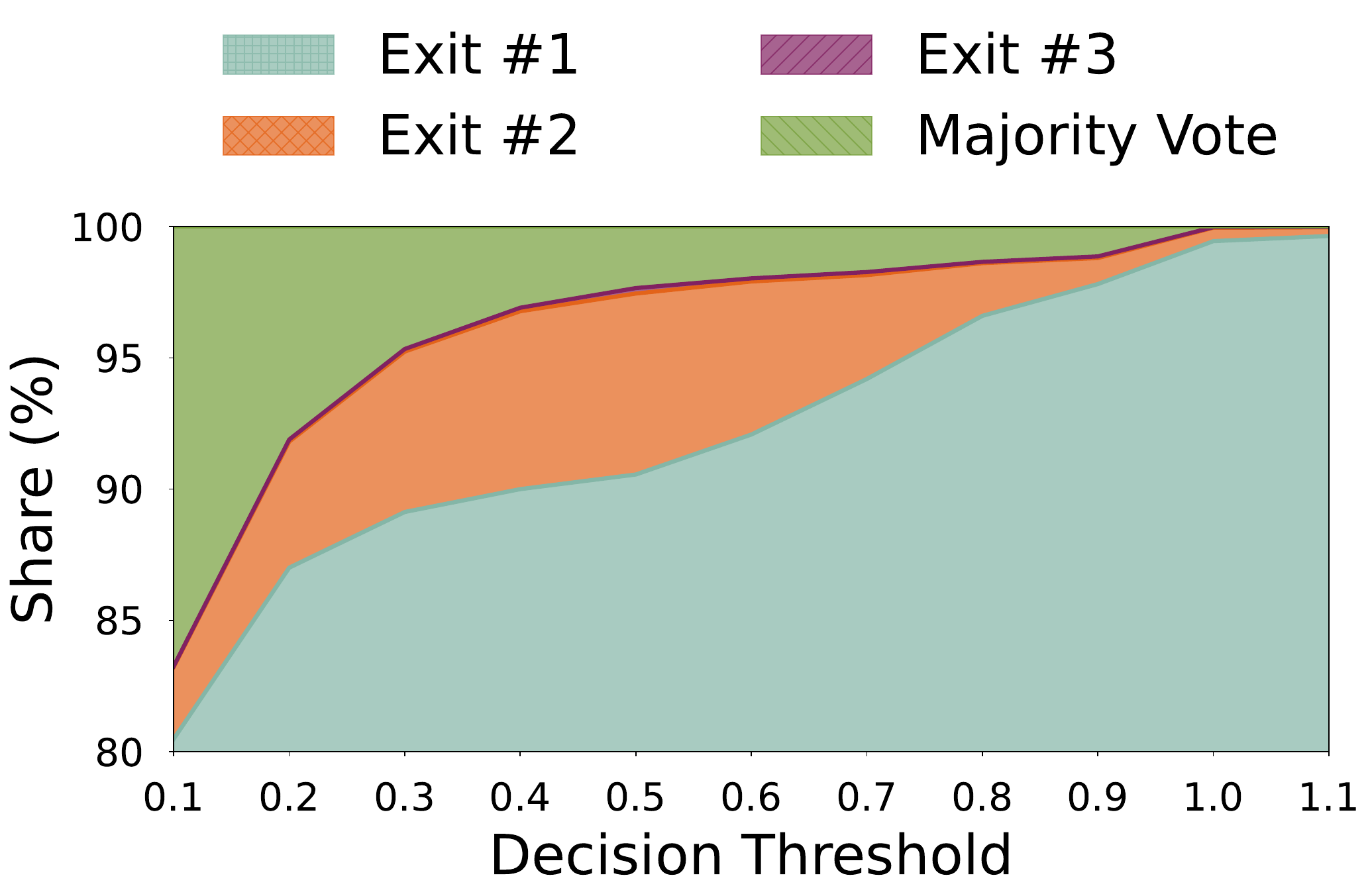}
\caption{Share of used classifiers for \ac{TP} \acp{EENN}.\\}
\label{fig:share_temp_patience}
\end{subfigure}
\caption{Benchmark results of a traditional Single Exit Neural Network, an optimized confidence-based \ac{EENN}, and our novel \ac{DD}- and \ac{TP}-\acp{EENN} across multiple detection thresholds.}
\end{figure*}

\section{Difference Detection Early Exit Neural Networks}

In existing \acp{EENN}, termination of the inference process is typically based on confidence-based metrics, which can lead to errors and vulnerabilities.

% TODO: improve the following sentences
To address this issue, we propose a \acl{DD} (\acs{DD})-based \ac{EENN}.
It leverages the temporal correlation within the sensor data and the propagation of changes through the network architecture to enable more efficient termination decisions.

The \ac{DD}-\ac{EENN} calculates the change in the classifier's output vector as the Euclidean distance between the current classification output vector ($\vec{o}$) and the vector of a previous sample ($\vec{o}_{t_{\text{initial}}}$).
This change is defined as the distance ($d$) between the output vectors of the first \ac{EE} classifier ($\vec{o}_{t,\text{exit}_0}$) between samples from two time-steps ($t_1$ and $t_{\text{initial}}$) (see Eq.~\ref{eq:change_metric}).

\begin{equation}
\small
\operatorname{change}(t_1, t_{\text{initial}}) = \operatorname{d}(\vec{o}_{t_1,\text{exit}_0},\vec{o}_{t_{\text{initial}},\text{exit}_0}) \quad , \quad \vec{o} \in \mathbb{R}^C
\label{eq:change_metric}
\end{equation}

We defined scenes as blocks of consecutive samples that are detected as similar by the decision mechanism.
The initial sample of a scene is labeled based on the majority vote of all classifiers, eliminating reliance on unreliable confidence-based metrics  (see Eq.~\ref{eq:majority_vote}, where $o_{t,\text{exit}} = \argmax(\vec{o}_{t,\text{exit}})$ and $n$ is the total number of classifiers in the \ac{EENN}).

\begin{equation}
\small
\operatorname{vote}({o_{t,\text{exit}_0}, o_{t,\text{exit}_1}, ..., o_{t,\text{exit}_n}}) = \argmax_{c \in C} \left(\sum_{i=1}^{n} [o_{t,\text{exit}_i} = c]\right)
\label{eq:majority_vote}
\end{equation}

The prediction at each time step is determined by comparing the currently processed sample to the initial sample.
If the change is smaller than the threshold, the prediction of the initial sample is reused, and no deeper layers and classifiers are executed.
If the change exceeds the threshold, the prediction is based on the majority vote of all classifiers, indicating the start of a new scene and setting the current sample as new reference (see Eq.~\ref{eq:update}).

\begin{equation}
\small
\operatorname{output}(t) = 
  \begin{cases}
    \operatorname{vote}_{t_{\text{initial}}} & \text{,if } \operatorname{change}(t, t_{\text{initial}}) < \mathrm{threshold} \\
    \operatorname{vote}_{t} & \text{,if } \operatorname{change}(t, t_{\text{initial}}) \ge \mathrm{threshold}
  \end{cases}
\label{eq:update}
\end{equation}

A similar threshold can be defined for regression tasks to compare the scalar output values of the compared time steps.

By comparing to the initial input of the scene rather than the direct predecessor, mislabeling due to slow drift in subsequent samples is prevented.
The use of the early classifier to calculate the change metric improves efficiency by reusing operations between the DD and the inference task.
Our approach provides a simple similarity measure for complex radar data without requiring domain knowledge, as in the case of template-matching solutions.
This similarity metric approach holds potential for other data modalities, which can be explored in future work.

\section{Temporal Patience Early Exit Neural Networks}

The \ac{DD}-\ac{EENN} relies on the similarity between samples to reuse previous predictions, making its efficiency and accuracy dependent on the defined threshold for an acceptable change.
However, always using the first \ac{EE} classifier in the network and not considering higher-level features extracted by deeper layers can lead to decreased accuracy.
To address this, we propose an improved solution with two modifications to enhance accuracy while leveraging the \ac{DD} approach.

The first adjustment involves the location of the \ac{DD} \ac{EE}-classifier in the network architecture.
Instead of always using the first classifier, this variant uses the first classifier that agrees with the majority vote of the initial sample of the current sequence for the following inputs of that sequence~(see Eq.~\ref{eq:tp:selection}).
This allows the mechanism to utilize a classifier more likely to detect a new scene accurately.

\begin{equation}
\small
\operatorname{select}(t) := \argmin_i ( o_{t,\text{exit}_i} = \operatorname{vote}(o_{t,\text{exit}_0}, o_{t,\text{exit}_1}, ..., o_{t,\text{exit}_n}) )
\label{eq:tp:selection}
\end{equation}

The change metric is updated to consider the selected classifier to accommodate this modification.
The modified change metric is shown in Eq.~\ref{eq:tp:change_metric}, where $\vec{o}$ represents the output vectors, $C$ is the number of classes, and $n$ is the selected classifier.

\begin{equation}
\small
\operatorname{change}(t, t_{\text{initial}}) = \operatorname{d}(\vec{o}_{t, \text{exit}_i},\vec{o}_{t_{\text{initial}},\text{exit}_i}) \quad , \quad \begin{aligned} \vec{o} \in \mathbb{R}^C \\ i = \operatorname{select}(t_{\text{initial}}) \end{aligned}
\label{eq:tp:change_metric}
\end{equation}

The second modification involves using the selected classifier to produce a label for subsequent samples.
This reduces reliance on the similarity threshold and adds minimal overhead as the \ac{DD}'s output vector is already calculated.

These modifications are intended to improve the prediction accuracy but introduce additional computations.
Scene change detection now considers the distance between the selected classifier's output vectors and the label change.
This approach incorporates a temporal component inspired by patience-based decision mechanisms.
The updated output function at each time step is described by Eq.~\ref{eq:tp:update}, where the condition determines if the change is within the threshold, and the output is based on the selected classifier or the majority vote.

\begin{equation}
\small
\begin{aligned}
\operatorname{condition}(t) &= \operatorname{change}(t, t_{\text{initial}}) < \mathrm{threshold} \quad \text{and} \\
&\quad\quad\argmax_{c \in C}(\vec{o}_{t,\text{exit}_i}) = \argmax_{c \in C}(\vec{o}_{t_\text{initial},\text{exit}_i}) \\
\operatorname{output}(t) &=
\begin{cases}
\vec{o}_{t,\text{exit}_i} & \text{, if } \operatorname{condition}(t) \\
\operatorname{vote}(o_{t,\text{exit}_0}, o_{t,\text{exit}_1}, \text{...}, o_{t,\text{exit}_n}) & \text{, otherwise}
\end{cases}
\end{aligned}
\label{eq:tp:update}
\end{equation}

\section{Benchmark}

To assess the performance of our method, we conducted a benchmark using a private radar dataset.
The used test set consists of 237,000 samples that belong to one of five classes as described in section~\ref{use_case}.

We compared our method to a confidence-based \ac{EENN} and a Single Exit version of the network architecture as these are the most commonly used methods.
The benchmark evaluated test set accuracy and inference efficiency for the people-counting task.
During the evaluation, multiple global threshold configurations were explored for our mechanisms, while the confidence-based \ac{EENN} utilized individual thresholds for each classifier.
To measure inference efficiency, we employed the mean of total \acl{MAC} (\acs{MAC}) operations per inference as a hardware-independent metric.

Additionally, we evaluated the majority voting mechanism against the state-of-the-art confidence-based labeling by applying both methods to the entire test set.

\subsection{Accuracy}

In our benchmark, we compared the \ac{EENN} decision methods and the Single Exit \ac{NN}.
The Single Exit network achieved the highest accuracy of 72.5\,\%.
The confidence-based \ac{EENN} achieved the second-highest accuracy of 71.5\,\%.
The \ac{DD} and \ac{TP} solutions achieved accuracies ranging from 69.71\,\% to 70.96\,\% and 71.12\,\% to 71.29\,\%, respectively.
Refer to Table ~\ref{tab:accuracy} for the performance of the individual classifiers and the final classifier.

While there was a slight decrease in accuracy, the drops were insignificant in practical applications.
The maximum accuracy drop observed was 2.79~\ac{p.p.} for the \ac{DD} approach and 1.38~\ac{p.p.} for the \ac{TP} approach. Compared to the confidence-based decision mechanism, the maximum accuracy drop was 1.79~\ac{p.p.} for \ac{DD} and 0.38~\ac{p.p.} for \ac{TP}.

The accuracy of the \ac{EENN} depends on the accuracy of individual classifiers and the at-runtime decision system. Future work could explore the calibration of exit-wise similarity thresholds to further improve accuracy.

\begin{table}[h]
\centering
\caption{The accuracy and computational cost of the classifiers of the \ac{EENN} model and the benchmark models.}
%\begin{tabular}{| c || c | c || c | c | c | c |}
\begin{tabularx}{1\columnwidth}{| X || X | X || X | X | X | X |}
\hline
 Model      & Single    & Conf.      & EE1      & EE2       & Final         & Maj.      \\ 
            & Exit      & EENN       &          &           & Class.        & Vote      \\
\hline
 Acc.       & 72.5\,\%  & 71.5\,\%   & 69.8\,\% & 71.3\,\%  & 72.5\,\%      & 71.5\,\% \\
 MACs       & 256.3k    & 216.2k     & 188.5k   & 222.2k    & 256.8k        & 308.8k    \\
\hline
%\end{tabular}
\end{tabularx}
\label{tab:accuracy}
\end{table}

This benchmark highlights the performance of the \ac{DD} and \ac{TP} mechanisms, and the detailed accuracies can be found in Fig.~\ref{fig:acc}.

\subsection{Efficiency}

The benchmark evaluated the mean cost in computations as \ac{MAC} operations per inference on the test set.
The Single Exit solution had the highest inference cost of 256\,kMAC per sample, while the confidence-based \ac{EENN} reduced it to 216.26\,kMAC  (see Fig.~\ref{fig:macs}).

The \ac{DD} and \ac{TP} mechanisms significantly reduced the mean inference cost, with both solution achieving a minimum cost of 188.6\,kMAC and a maximum cost of 209.7\,kMAC.
The average cost across all threshold configurations for the \ac{DD} was 192.8\,kMAC and 193.5\,kMAC
for the \ac{TP} solution.
Compared to the confidence-based solution, these approaches offered inference cost reductions of up to 26\,\%.
The efficiency improvement of the \ac{DD} and \ac{TP} solutions came from their execution strategy, where only the earliest or previously selected exit classifier was executed as long as the change between the current and initial samples remained below the threshold.
In contrast, confidence-based methods executed all exit calculations until a sufficiently confident result was obtained.

While the \ac{TP} solution incurred additional costs for a small number of subsequent samples, it showed improved classification capabilities (see Fig.~\ref{fig:share_diff_detect} and~\ref{fig:share_temp_patience}).
The \ac{DD} and \ac{TP} mechanisms demonstrated superior efficiency compared to the confidence-based approach.

\subsection{New Scene Labeling}

The \ac{DD}- and \ac{TP}-\acp{EENN} employ a majority voting mechanism for labeling the initial frame of newly detected scenes.
This was intended to prevent overthinking - a process in \acp{EENN} where \acp{EE} produce a correct prediction but are overwritten by deeper classifiers~\cite{kayaShallowdeepNetworksUnderstanding2019}.
Comparing the accuracy of the majority voting approach to the confidence-based solution, we found that the majority vote achieved an accuracy of 71.54\,\%, while the confidence-based approach achieved an accuracy of 71.42\,\%.
The difference between the two approaches was only 0.16 \ac{p.p.}, indicating their similarity in accuracy for this application and that overthinking is not an issue for this application.

Interestingly, we observed that an optimally tuned confidence-based \ac{EENN} with exit-wise thresholds can compete with the majority vote mechanism in terms of accuracy.
This suggests that both approaches have their strengths and weaknesses, and the choice between them may depend on the specific requirements of the task and dataset.

\section{Conclusion}

We have introduced a novel approach for processing radar data using \acp{EENN} with temporal information.
Our evaluation demonstrated significant efficiency gains of up to 26\,\% compared to the Single Exit network.
%By combining the novel decision mechanisms with data reordering preprocessing, we achieved an overall efficiency gain of 70\,\%-75\,\% over the original architecture that relied on temporal 3D convolutions.
Both the \ac{DD}-\ac{EENN} and \ac{TP}-\ac{EENN} solutions are easy to implement and provide cost-effective similarity measurements through computation reuse.
While the \ac{DD}-\ac{EENN} achieved slightly higher efficiency gains, the \ac{TP}-\ac{EENN} maintained higher accuracy across the test set and is less dependent on the threshold hyperparameter configuration.
The majority vote and the optimally tuned confidence-based \ac{EENN} yielded similar accuracies for this use case, indicating the potential for even greater efficiency gains when combining confidence-based labeling for new scenes with the \ac{DD} approach.
Our solution does not require specialized hardware and can be combined with additional optimizations such as quantization and pruning.
%Future work can explore attack and error mitigation or generalize the solution to other data modalities.
Future work can focus on utilizing the similarity measurement for additional use cases like monitoring or could evaluate the approach on other data modalities like audio or image data exploring various \ac{EENN} architectures.

We have shown the possibility of leveraging the temporal correlation of sensor data for more efficient termination decisions on \acp{EENN}.
This research provides a strong foundation for future work and has the potential to impact various applications.
%\clearpage
%%%%%%%%%%%%%%%%%%%%%%%%%%%%%%%%%%%%%%%%%%%%%%%%%%%%%%%%%%%%%%%%
%                        APPENDIX                              %
%%%%%%%%%%%%%%%%%%%%%%%%%%%%%%%%%%%%%%%%%%%%%%%%%%%%%%%%%%%%%%%%

% \section*{Acknowledgment}
% The project “RadarSkin” has received funding from the \ac{BMBF} under the call “Electronic Systems for Edge Computing” (grant number 16ME0543). The responsibility for the content of this publication lies with the author.

\bibliographystyle{IEEEtran}
\bibliography{een, radar}

\end{document}